\documentclass[12pt]{article}

\usepackage[ruled,vlined]{algorithm2e}
\usepackage{scalerel}

\usepackage{amssymb}
\usepackage{multirow}
\usepackage{tabularx}
\usepackage{amsmath}
\usepackage{hyperref}
\usepackage{graphicx}
\usepackage{url}
\usepackage[detect-all]{siunitx}
\usepackage{booktabs}
\usepackage{bm,bbm}
\usepackage{setspace}
\usepackage[caption=false, font=footnotesize, margin=.5cm]{subfig}
\usepackage{float}
\usepackage[noabbrev]{cleveref}
\usepackage{xcolor}
\usepackage{soul}
\usepackage{cases}
\usepackage{orcidlink}
\usepackage{MnSymbol}
\usepackage{mathtools}
\usepackage{color}

\usepackage{microtype}
\usepackage{rotating}

\usepackage{arydshln}
\usepackage{pgfplots}
    \pgfplotsset{compat=1.18} 
\usepackage{xparse}
\NewDocumentCommand\angRange{O{} m m}{\SIrange[parse-numbers=false, #1]{\ang[parse-numbers=true]{#2}}{\ang[parse-numbers=true]{#3}}{}}

\usepackage[flushleft]{threeparttable}

\usetikzlibrary{shapes.geometric, arrows}

\tikzstyle{startstop} = [rectangle, rounded corners, minimum width = 3cm, minimum height = 1cm, text centered, draw = black, fill = red!30]
\tikzstyle{io} = [trapezium, trapezium left angle = 70, trapezium right angle = 110, minimum width = 3cm, minimum height = 1cm, text centered, draw = black, fill = blue!30]
\tikzstyle{process} = [rectangle, minimum width = 3cm, minimum height = 1cm, text centered, draw = black, fill = orange!30]
\tikzstyle{decision} = [diamond, minimum width = 3cm, minimum height = 1cm, text centered, draw = black, fill = green!30]
\tikzstyle{arrow} = [thick,->,>=stealth]

\usepackage{graphbox}

\usepackage[noadjust]{cite}

\usepgfplotslibrary{colormaps}
\pgfplotsset{
    colormap={cmaptheta}{
        rgb255=(205.65,255,41.13)
        rgb255=(123.06,255,123.71)
        rgb255=(40.48,255,206.29)
        rgb255=(0,178.9,255)
        rgb255=(0,0,255)
    }
}

\definecolor{gr}{HTML}{00ff00} 
\definecolor{cy}{HTML}{00ffff} 
\definecolor{re}{HTML}{FF0000} 
\definecolor{bl}{HTML}{0000ff} 
\definecolor{mg}{HTML}{00ff00} 

\definecolor{mycolor}{rgb}{0.17, 0.84, 0.84}  

\graphicspath{{}}

\title{ArcGate: Adaptive Arctangent Gated Activation}

\author{
Avik Bhattacharya\orcidlink{0000-0001-6720-6108}\\
Microwave Remote Sensing Lab\\
Center of Studies in Resources Engineering\\
Indian Institute of Technology Bombay, India\\
\texttt{avikb@csre.iitb.ac.in}
\and
Siddhant Dnyanesh Gole\orcidlink{0009-0008-0958-4109}\\
Centre of Machine Intelligence and Data Science\\
Indian Institute of Technology Bombay, India\\
\texttt{siddhant.gole@iitb.ac.in}
\and
Subhasis Chaudhuri\orcidlink{0000-0002-1680-0016}\\
Department of Electrical Engineering\\
Indian Institute of Technology Bombay, India\\
\texttt{sc@ee.iitb.ac.in}
\and
Alejandro C.\ Frery\orcidlink{0000-0002-8002-5341}\\
School of Mathematics and Statistics\\
Victoria University of Wellington\\
Wellington 6140, New Zealand\\
\texttt{alejandro.frery@vuw.ac.nz}
\and
Biplab Banerjee\orcidlink{0000-0001-8371-8138}\\
Center of Studies in Resources Engineering\\
Indian Institute of Technology Bombay, India\\
\texttt{bbanerjee@iitb.ac.in}}

\date{}

\begin{document}

\maketitle

\begin{abstract}

Activation functions are central to deep networks, influencing non-linearity, feature learning, convergence, and robustness. 
This paper proposes the Adaptive Arctangent Gated Activation (ArcGate) function, a flexible formulation that generates a broad spectrum of activation shapes via a three-stage non-linear transformation. 
Unlike conventional fixed-shape activations such as ReLU, GELU, or SiLU, ArcGate uses seven learnable parameters per layer, allowing the neural network to autonomously optimize its non-linearity to the specific requirements of the feature hierarchy and data distribution. 
We evaluate ArcGate using ResNet-50 and Vision Transformer (ViT-B/16) architectures on three widely used remote sensing benchmarks: PatternNet, UC Merced Land Use, and the 13-band EuroSAT MSI multispectral dataset. 
Experimental results show that ArcGate consistently outperforms standard baselines, achieving a peak overall accuracy of \SI{99.67}{\percent} on PatternNet. 
Most notably, ArcGate exhibits superior structural resilience in noisy environments, maintaining a \SI{26.65}{\percent} performance lead over ReLU under moderate Gaussian noise ($\bm{\sigma=0.1}$). 
Analysis of the learned parameters reveals a depth-dependent functional evolution, where the model increases gating strength in deeper layers to enhance signal propagation. 
These findings suggest that ArcGate is a robust and adaptive general node activation function
for high-resolution earth observation tasks.
\end{abstract}

\bigskip


\section{Introduction}
\label{sec:intro_existing_target}

Activation functions provide the necessary non-linearity that enables deep neural networks to learn and approximate complex relationships between inputs and outputs. 
Historically, saturating activation functions such as the sigmoid~\cite{Mitchell1997}, $\sigma(x)=(1 + e^{-x})^{-1}$, and the hyperbolic tangent~\cite{Hu2023-vn}, $\tanh(x)$, were the standard choices. 
Although these functions produce bounded outputs, they are prone to the vanishing gradient problem in deep architectures, especially when inputs fall within the saturation regions.

To alleviate the gradient dissipation problem, the Rectified Linear Unit (ReLU)~\cite{Dey2021-lq}, defined as $\text{ReLU}(x) = \max\{0, x\}$, emerged as the predominant activation function in convolutional neural networks (CNNs). 
ReLU is computationally efficient and enables sparse activations; however, it suffers from the \emph{dying ReLU} problem, in which neurons may become inactive if their inputs remain consistently negative, resulting in zero gradients. 
Variants such as Leaky ReLU~\cite{Palsson2021-gf} and Parametric ReLU (PReLU)~\cite{Luo2025-xf} mitigate this issue by assigning a small, and in the case of PReLU, learnable slope to negative input values.

Recent developments have shifted attention toward smooth, non-monotonic activation functions. 
The Sigmoid Linear Unit (SiLU)~\cite{Chatterjee2025-oo}, also widely known as the Swish function, is defined as the input multiplied by its sigmoid, $\text{Swish}(x) = x \, \sigma(x)$. 
Similarly, the Gaussian Error Linear Unit (GELU)~\cite{Zhong2025-kx} is defined as $ \text{GELU} (x) = x \, \Phi(x)$, where $\Phi(x)$ denotes the standard Gaussian cumulative distribution function. 
Both functions have demonstrated superior performance in transformer-based architectures and deep CNNs. 
By allowing a controlled range of negative outputs and possessing smooth, continuous derivatives, they facilitate more stable gradient flow and improved optimization. 

Despite these advances, those activation functions do not adapt to the problem specifics. 
This lack of structural flexibility motivates the development of an adaptive parametric formulation that can shape itself to the requirements of different network layers and data distributions. 

\section{The Adaptive Activation Function}
\label{sec:Methodology}

We introduce a node adaptive parametric formulation based on a sequence of nonlinear transformations: a monotonic transition, a non-linear stretching operation, and an affine non-linear combination.
These three components produce a flexible family of activation functions expressed as:
\begin{align}
    u(x; a,c) &= \frac{1}{2} + \frac{1}{\pi}\tan^{-1}\big[a(x - c)\big], \\
    v(x ; p) &= \frac{2}{\pi}\tan^{-1}\!\Bigg[\frac{u(x; a,c)}{1 - u(x; a,c)}\Bigg]^{p} , \\    
    F(x ; \alpha, \beta, \gamma, \delta) &= (\alpha\, x + \beta)\, v(x; p) + (\gamma\, x + \delta),
\end{align}
where $a,p \in\mathbbm R_{>0}$ are shape parameters, $x,c,\alpha,\beta,\delta,\gamma\in\mathbbm R$, making $u,v,\in (0,1)$ and $F \in\mathbbm R$. 
Although we denoted these parameters as fixed, they are learned during training.
The following subsections provide a detailed description of each component of the adaptive activation function.

\subsection{Monotonic Transition Function}

The first stage defines a smooth, monotonic transition as:
\begin{equation}
    u(x; a,c) = \frac{1}{2} + \frac{1}{\pi}\tan^{-1}\big[a(x - c)\big].
\end{equation}
The function $u(x; a,c)$ smoothly maps $x\in\mathbbm R$ to $(0,1)$, where $a$ controls steepness and $c$ sets the transition center. Its arctangent form ensures smoothness and numerical stability, while the symmetry 
\begin{equation}
    u(x; -a,c) = 1 - u(x; a,c).
\end{equation}
shows that changing the sign of $a$ reverses the transition direction.

\subsection{Nonlinear Stretching}

The second stage modifies the transition by introducing an adaptive smooth gating function as:
\begin{equation}
 v(x; p) = \frac{2}{\pi}\tan^{-1}\!\Bigg[\frac{u(x; a,c)}{1 - u(x; a,c)}\Bigg]^{p}.
\end{equation}
The ratio ${u}({1-u})^{-1}$ performs an odds transformation, mapping $(0,1)$ to $(0,\infty)$. The exponent $p$ controls the gate sharpness: larger values produce steeper, switch-like transitions, while smaller values yield smoother responses. The arctangent then maps the result back to $(0,1)$, preserving boundedness. Consequently, $v(x; p)$ forms a smooth adaptive gating family, ranging from sigmoid-like to near piecewise-linear behavior. Similar to the transition function, $v(x; p)$ also satisfies the symmetry property as:
\begin{equation}
    v(x; -p) = 1 - v(x; p).
\end{equation}

\subsection{Adaptive Arctangent Gated Activation Function}

The final Adaptive Arctangent Gated Activation (ArcGate) function is obtained by combining the gating function $v(x;p)$ with linear and bilinear terms, yielding an affine mapping:
\begin{equation}
F(x ; \alpha, \beta, \gamma, \delta) = (\alpha\, x + \beta)\, v(x; p) + (\gamma\, x + \delta),
\end{equation}
Each parameter plays a distinct role in shaping the activation function:
\begin{itemize}
    \item $\alpha$ governs the interaction between the input and the non-linear component, effectively regulating the slope in the active region,
    \item $\beta$ controls the relative contribution of the non-linear term,
    \item $\gamma$ introduces a purely linear component, and
    \item $\delta$ provides a constant shift.
\end{itemize}

Together, these components give the proposed formulation high flexibility and expressive power. 
By appropriately tuning the parameters ${a, c, p, \alpha, \beta, \gamma, \delta}$, the model represents a wide range of activation behaviors, including linear, saturating, rectifying, and smoothly varying nonlinear functions. 
Consequently, many classical and modern activation functions arise as special cases within this unified, adaptive parametric framework.

\section{Mathematical Properties and Generalization}

ArcGate forms a functional superset of a broad class of activation functions. 
By adapting the parameter vector $(a, c, p, \alpha, \beta, \gamma, \delta)$, it can smoothly traverse a continuous functional space spanning multiple activation behaviors and transition seamlessly between distinct activation families.

\subsection{Functional Generalization}
The power of ArcGate lies in its ability to replicate classical and state-of-the-art activations as special cases or close approximations. 

\begin{itemize}
    \item \textbf{Rectification and Gating:} By setting $\alpha=1$ and $\beta, \gamma, \delta = 0$, ArcGate operates in a gating mode where $F(x) = x \, v(x)$. 
	In the limit $a, p \to \infty$, the function $v(x)$ converges to a Heaviside step function $H(x-c)$, thereby replicating the ReLU activation.
    \item \textbf{Saturating Functions:} By setting $\alpha=0$ and $\beta=1$, the function discards the linear input interaction and becomes sigmoidal. 
    With moderate values of $a$ and $p=1$, ArcGate approximates the Sigmoid function. 
    Furthermore, hyperbolic tangent (tanh) is realized by scaling and shifting via $\beta=2$ and $\delta=-1$.
    \item \textbf{Smooth Non-monotonicity:} Modern activations like SiLU (Swish) and GELU emerge when $\alpha=1$ and the $a$ and $p$ are tuned to provide specific curvatures. 
    The differentiability of the arctangent function ensures that these profiles remain smooth across the domain.
    \item \textbf{Linearity and Leakage:} The parameter $\gamma$ provides a direct linear path. 
When added to the gating mode, ArcGate replicates the Leaky~ReLU behavior, ensuring a non-zero gradient for negative inputs. 
If $\alpha, \beta, \delta = 0$ and $\gamma=1$, ArcGate collapses into the identity function.
\end{itemize}

\subsection{Analytical Properties}
Beyond generalization, the mathematical structure of ArcGate offers several advantages for deep learning optimization:

\subsubsection{Continuity and Differentiability}
Unlike ReLU or PReLU, which possess a non-differentiable ``kink'' at the origin, ArcGate is a composition of arctangent and power functions. 
It is therefore $C^{\infty}$ (infinitely differentiable) for all $x \in \mathbbm{R}$. 
This smoothness prevents the ``jitter'' in gradient updates and facilitates more stable backpropagation, which is critical for the 50-layer architectures used in this study.

\subsubsection{Numerical Stability}
Using the arctangent function as the primary squashing mechanism ensures that the internal components $u(x)$ and $v(x)$ are bounded to $(0, 1)$. 
Unlike the exponential terms in Sigmoid or Swish, which can lead to overflow in half-precision training, the arctangent remains numerically stable even for very large input magnitudes.

\subsubsection{Learnable Bias and Slope}
The parameters $c$ and $\delta$ allow the activation function to learn an internal bias, independent of the weight layers. 
This dual-bias mechanism enables the network to center the activation threshold ($c$) and the output baseline ($\delta$) to match the specific statistics of multi-spectral or SAR data distributions, which often exhibit high dynamic ranges.

\section{Results and Discussion}
\label{sec:results_discussion}

\subsection{Implementation Details}
Experimental evaluations were performed using both the ResNet-50~\cite{resnet} and Vision Transformer (ViT-B/16)~\cite{vit} architectures. 
For the ResNet-50 backbone, all standard ReLU layers were replaced with the proposed learnable ArcGate modules. 
Similarly, for the ViT-B/16 architecture, the standard GELU activations were replaced with the proposed ArcGate modules. 
The ArcGate parameters $(a, c, p, \alpha, \beta, \gamma, \delta)$ were initialized to $(5.0, 0.0, 1.0, 1.0, 0.0, 0.0, 0.0)$, corresponding to a Soft~ReLU configuration that enables stable gradient propagation during early training.

Experiments were conducted on PatternNet~\cite{patternnet}, UC Merced~\cite{ucmerced}, and 13-band EuroSAT~\cite{eurosat}. For EuroSAT, ResNet-50 (\texttt{conv1}) and ViT-B/16 (\texttt{conv\_proj}) were adapted to 13-channel inputs, with the extra 10 bands initialized to the mean of the pre-trained RGB filters to preserve ImageNet priors. All models were trained for 50 epochs using AdamW ($10^{-4}$ learning rate, $10^{-2}$ weight decay). Inputs were resized to $224 \times 224$, augmented with random flips and cropping, and multispectral reflectance values were scaled by \num{10000}. Training was performed on an NVIDIA L40S GPU. 

\subsection{Comparative Analysis}

ArcGate was benchmarked against standard and state-of-the-art activations (Table~\ref{tab:results}) and consistently outperformed all baselines, reaching \SI{94.72}{\percent} on UC Merced (\SI{1.67}{\percent} above SiLU) and \SI{96.25}{\percent} on PatternNet. These results show that its learnable parametric form adapts better to diverse spatial patterns than fixed activations. By optimizing the nonlinear response at each layer, ArcGate is particularly effective for high-resolution land-cover classification, where activation behavior differs between shallow and deep layers.

\begin{sidewaystable}[htbp]
\centering
\caption{Overall Accuracy (OA \si{\percent}) Comparison Across RGB and Multispectral Datasets}
\label{tab:results}
\begin{tabular}{lcccccccc}
\toprule
 & \multicolumn{2}{c}{\textbf{UC Merced}} & & \multicolumn{2}{c}{\textbf{PatternNet}} & & \multicolumn{2}{c}{\textbf{EuroSAT MSI}} \\ \cline{2-3} \cline{5-6} \cline{8-9} 
\textbf{Activation Function} & \textbf{ResNet-50} & \textbf{ViT-B/16} & & \textbf{ResNet-50} & \textbf{ViT-B/16} & & \textbf{ResNet-50} & \textbf{ViT-B/16} \\ \midrule
Sigmoid      & 14.04 & 75.00 & & 16.63 & 95.61 & & 62.28 & 94.69 \\
Tanh         & 90.95 & 55.00 & & 98.27 & 92.89 & & 96.67 & 91.94 \\
ReLU         & 97.85 & 96.90 & & 99.57 & 99.19 & & \textbf{98.52} & 96.01 \\
Leaky-ReLU   & 98.09 & 97.38 & & 99.57 & 99.18 & & 98.43 & 95.30 \\
SiLU (Swish) & 91.90 & 95.47 & & 99.01 & 96.99 & & 97.56 & 94.59 \\
GELU         & 96.19 & \textbf{99.52} & & 99.18 & 99.39 & & 98.06 & \textbf{97.90} \\ \midrule
\textbf{ArcGate (Ours)} & \textbf{98.15} & 98.33 & & \textbf{99.67} & \textbf{99.48} & & 98.47 & 97.69 \\
\bottomrule
\end{tabular}%
\end{sidewaystable}

\subsection{Noise Robustness and Structural Resilience}
To evaluate robustness to signal perturbations, which are common in remote sensing due to atmospheric disturbances and sensor noise, we performed a stress test adding Gaussian noise with standard deviations ranging from $0$ (noise-free) to $0.2$ to the input images.

We evaluated two models on the PatternNet validation set: a baseline model with a natively optimized ReLU activation and a model incorporating the proposed ArcGate. 
Table~\ref{tab:noise_results} shows the classification performance at each noise level.

\begin{table}[htbp]
\centering
\caption{Robustness Comparison under Gaussian Noise ($\sigma$) on PatternNet dataset.}
\label{tab:noise_results}
\resizebox{0.75\columnwidth}{!}{%
\begin{tabular}{lccccc}
\toprule
\textbf{Model} & \textbf{0.0} & \textbf{0.05} & \textbf{0.10} & \textbf{0.15} & \textbf{0.20} \\ \midrule
ReLU & 99.57 & 96.61 & 39.34 & 18.06 & 12.60 \\
\textbf{ArcGate} & \textbf{99.67} & \textbf{98.98} & \textbf{65.99} & \textbf{31.18} & \textbf{18.11} \\ \midrule
\textbf{Gain ($\Delta$)} & +0.10 & +2.37 & \textbf{+26.65} & +13.12 & +5.51 \\ 
\bottomrule
\end{tabular}
}
\end{table}

Without noise ($\sigma = 0.0$), both models achieve strong performance, with ArcGate outperforming the ReLU baseline (\SI{99.67}{\percent} versus \SI{99.57}{\percent}). 
As noise levels increase, a clear performance gap emerges. 
At a moderate noise level ($\sigma = 0.10$), the ReLU-based model suffers severe accuracy degradation, dropping to \SI{39.34}{\percent}. 
In contrast, the ArcGate-enabled model remains substantially more stable, achieving an accuracy of \SI{65.99}{\percent}. 
This corresponds to a \SI{26.65}{\percent} absolute improvement over the ReLU baseline.

This robustness stems from ArcGate's learned geometric profile. 
Unlike ReLU's sharp, non-differentiable discontinuity at the origin, ArcGate adapts its parameters ($a,p$) to form a smooth, fully differentiable transition that naturally regularizes the response. 
This smoothness reduces sensitivity to small perturbations and limits noise amplification, making ArcGate a more stable and reliable alternative for Earth observation tasks with variable data quality.

\subsection{Layer-wise Functional Adaptation}
\begin{figure}[htbp]
    \centering
    \includegraphics[width=0.8\columnwidth]{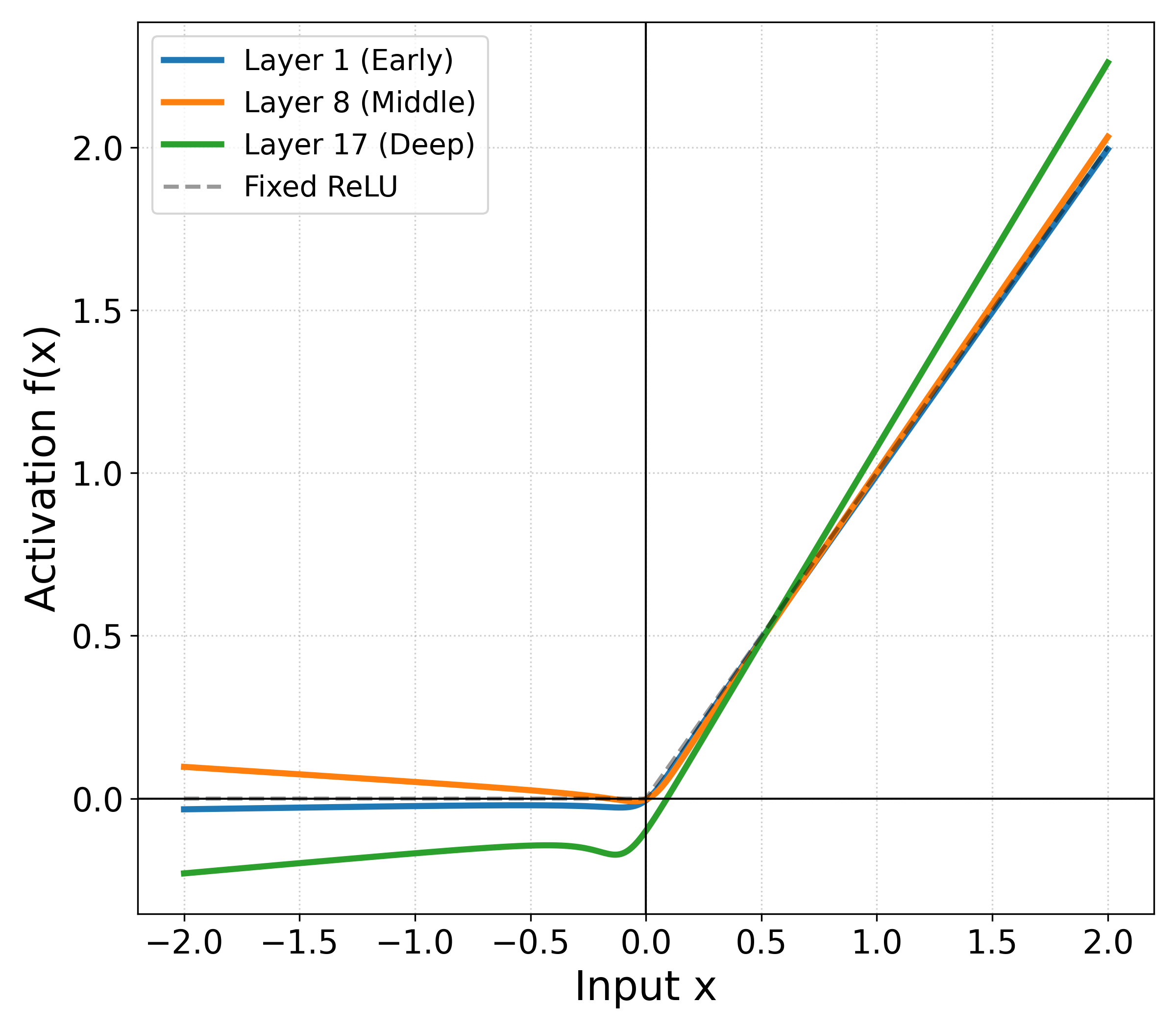}
    \caption{Depth-wise adaptation of ArcGate in ResNet-50, with increasing gain ($\alpha$) in deeper layers to counter attenuation, unlike fixed ReLU (dashed).}
    \label{fig:evolution}
\end{figure}

To verify the hypothesis that optimal activation behavior is depth-dependent, we analyzed the learned ArcGate parameters across ResNet-50 after 50 epochs of training on PatternNet. Fig.~\ref{fig:evolution} shows the activation curves for an early layer (L1), a middle layer (L25), and a deep layer (L49), compared with the standard ReLU baseline.

The results reveal a clear hierarchical adaptation of the activation function. All layers preserve a smooth Soft-ReLU transition near the origin, mitigating the gradient discontinuity of ReLU, while the gating slope ($\alpha$) progressively increases with depth. Specifically, $\alpha$ evolves from 0.99 in shallow layers to 1.12 in the deepest layers.

This behavior indicates that the network autonomously learned to amplify signal intensity in deeper layers. 
In high-resolution remote sensing tasks, this adaptation is critical for two reasons:
\begin{enumerate}
    \item \textbf{Signal Preservation:} Increasing the slope in deeper layers helps counteract potential signal attenuation (vanishing gradients) over the 50-layer architecture.
    \item \textbf{Feature Saliency:} Deeper layers responsible for complex semantic feature extraction require higher representational gain to distinguish between subtle inter-class variances in land-cover patterns.
\end{enumerate}

The fact that ArcGate converges to these distinct, layer-specific profiles confirms that a \emph{one-size-fits-all} activation, such as fixed ReLU, is suboptimal. 
The parametric flexibility of ArcGate enables a self-organizing non-linearity that adapts to the specific requirements of the feature hierarchy, contributing to the superior performance and robustness observed in the earlier experiments.

\begin{figure*}[htbp]
    \centering
    \subfloat[Steepness]{%
    \includegraphics[width=0.32\textwidth]{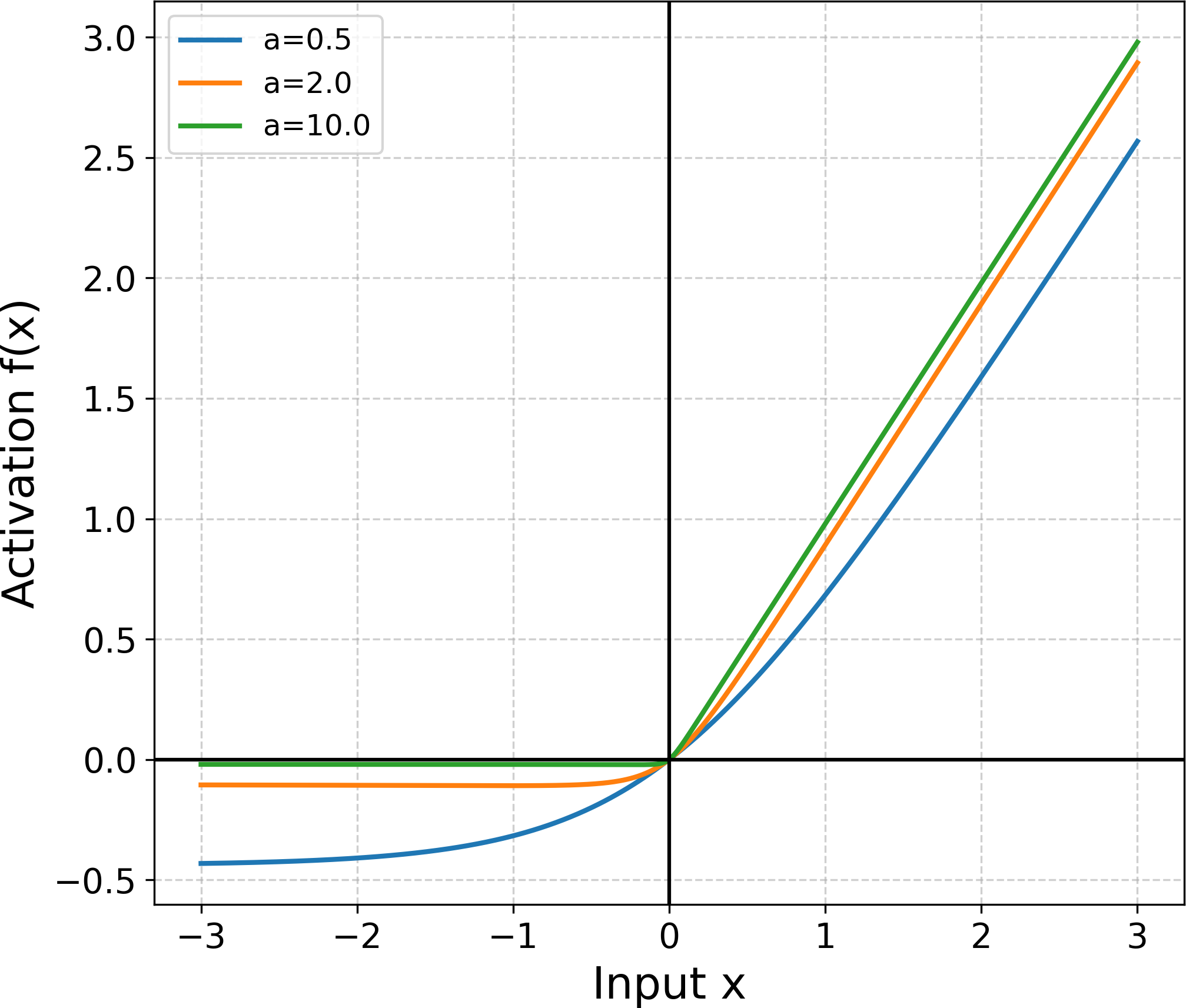}}
    \hfill
    \subfloat[Sharpness]{%
    \includegraphics[width=0.32\textwidth]{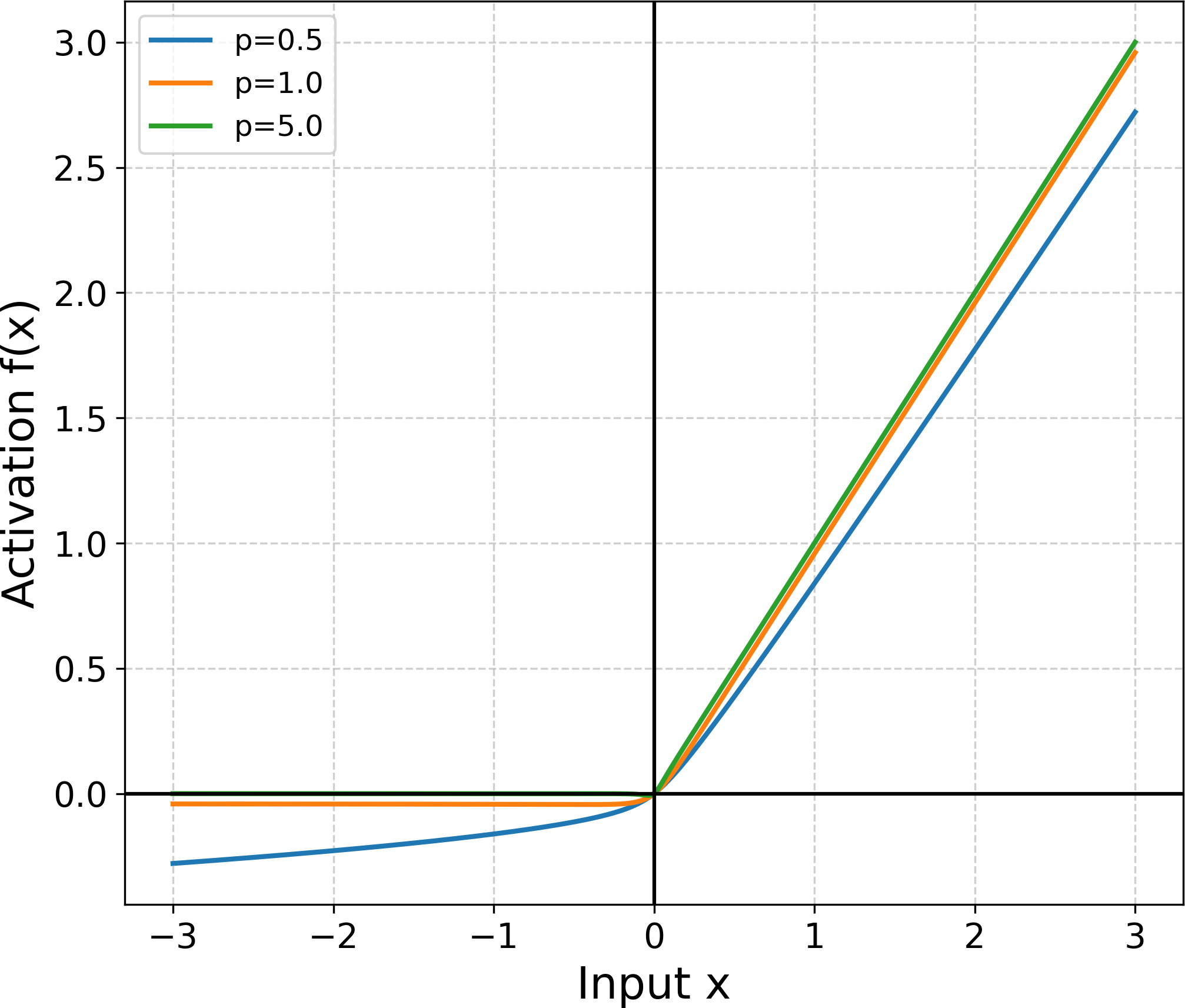}}
    \hfill
    \subfloat[Horizontal shift]{%
    \includegraphics[width=0.32\textwidth]{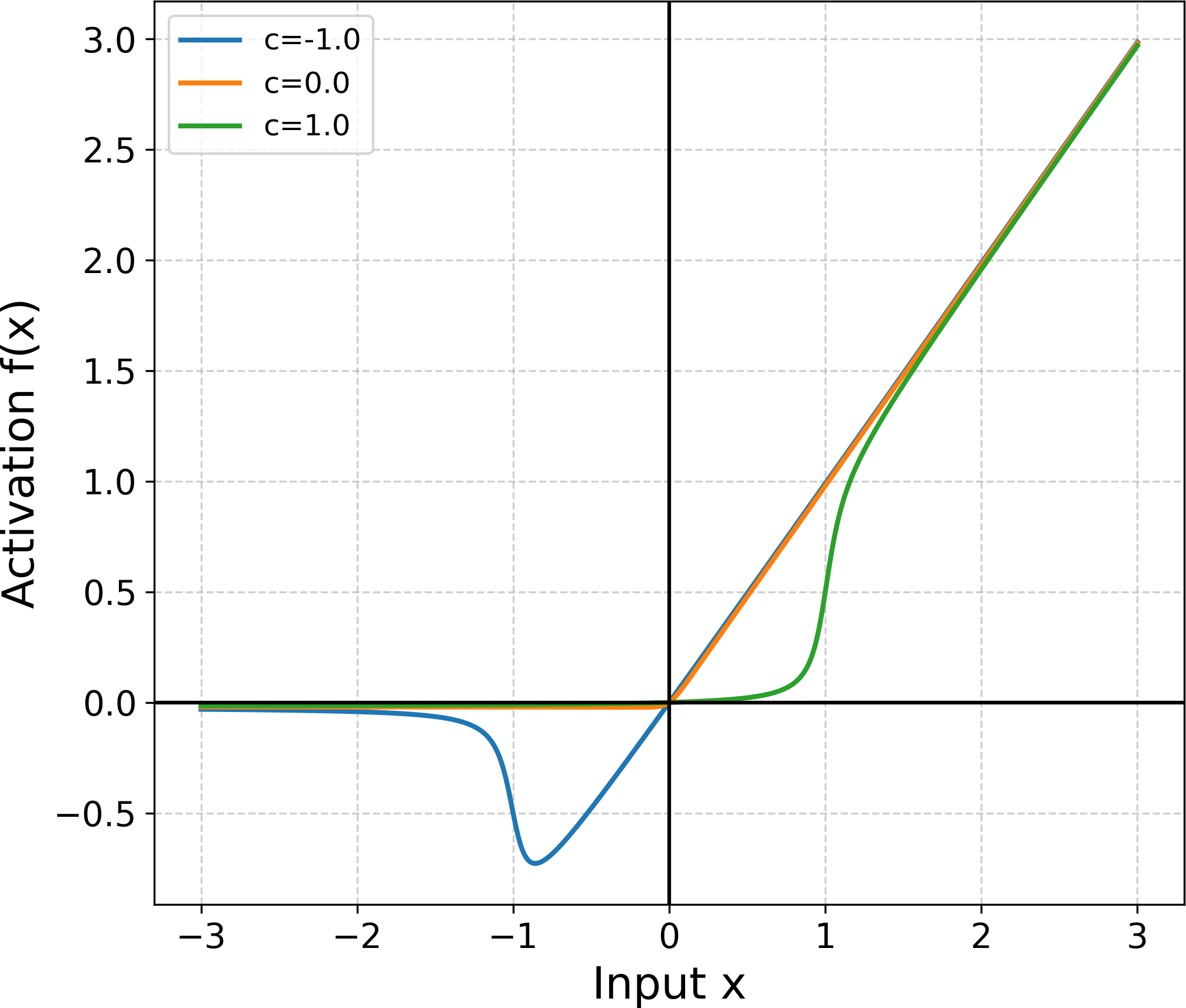}}
    \\
    \subfloat[Function class]{%
    \includegraphics[width=0.32\textwidth]{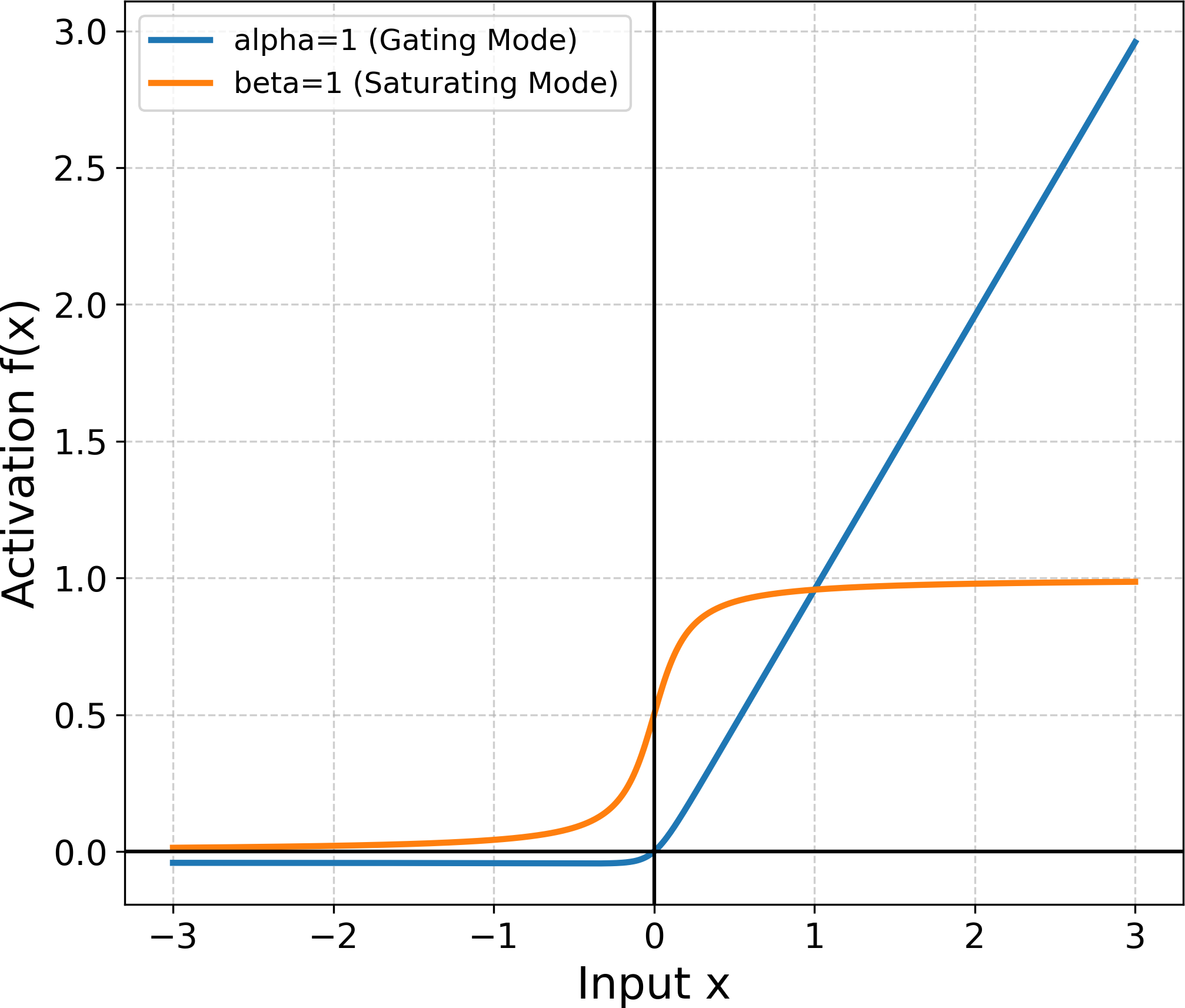}}
    \hfill
    \subfloat[Linear trend]{%
    \includegraphics[width=0.32\textwidth]{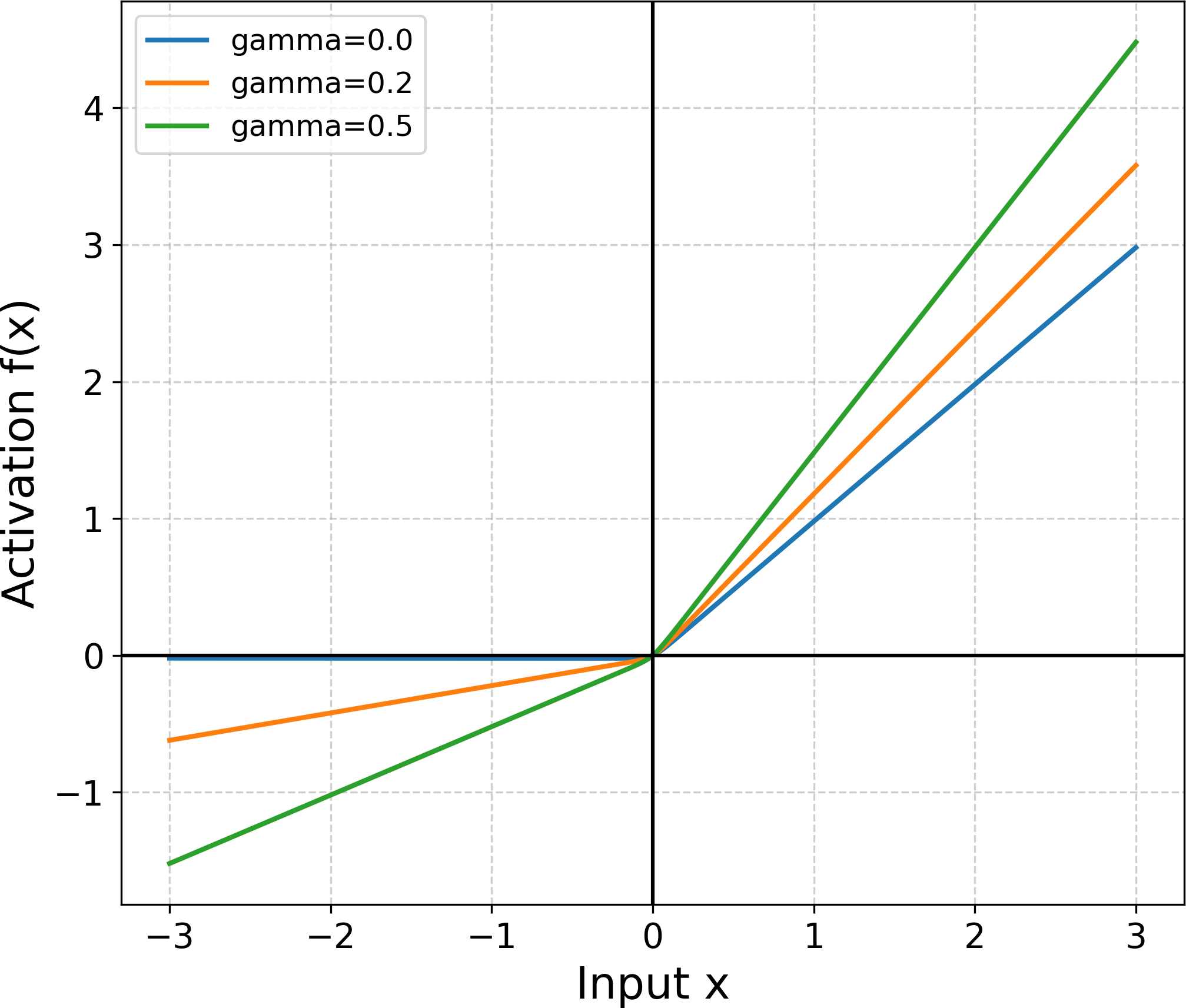}}
    \hfill
    \subfloat[Classical functions]{%
    \includegraphics[width=0.32\textwidth]{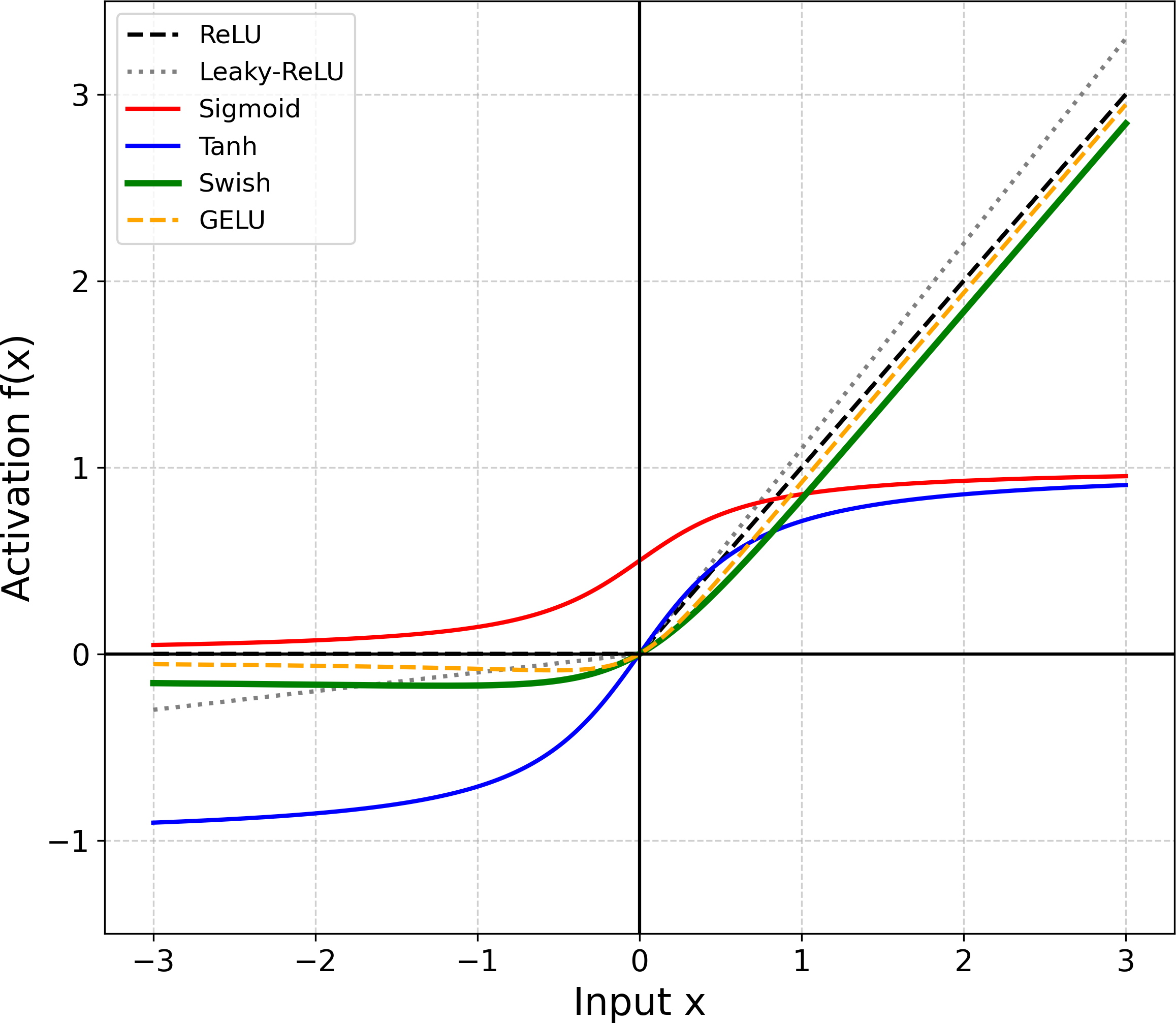}}
    \caption{Parametric sensitivity analysis of ArcGate: 
    (a)~influence of steepness $a$ on the transition slope, 
    (b)~effect of sharpness $p$ on corner curvature, 
    (c)~horizontal translation via the $c$ parameter, 
    (d)~distinction between gating ($\alpha$) and saturating ($\beta$) modes, 
    (e)~introduction of linear leakage via $\gamma$, 
    and (f)~replication of classical activation functions within the adaptive framework.}
    \label{fig:sensitivity}
\end{figure*}

\subsection{Ablation Study: Influence of Initialization}

On the PatternNet dataset, ArcGate initialization was evaluated using Soft-ReLU, Identity, Random, and a ReLU baseline (Table~\ref{tab:ablation_init}). All variants converged, but initialization significantly affected performance. Soft-ReLU provided stable, near-optimal results with faster convergence and accuracy matching or exceeding ReLU, whereas random initialization was slower. This confirms the value of physically motivated initialization for robust remote sensing applications.

\begin{table}[htbp]
\centering
\caption{Initialization Strategies on PatternNet Dataset}
\label{tab:ablation_init}
\resizebox{0.75\columnwidth}{!}{%
\begin{tabular}{lcc}
\toprule
\textbf{Init. Strategy} & \textbf{ResNet-50} & \textbf{ViT-B/16} \\ \midrule
ReLU (Baseline)         & 99.57              & 99.19             \\
ArcGate (Identity)         & 98.42              & 96.85             \\
ArcGate (Random)           & 97.15              & 94.22             \\
\textbf{ArcGate (Soft-ReLU)} & \textbf{99.67}   & \textbf{99.08}    \\ \bottomrule
\end{tabular}
}
\end{table}
\subsection{Ablation Study: Parametric Complexity vs. Performance}
To justify the structural design of ArcGate, we conducted an ablation study on the PatternNet dataset using three configurations: 
(i)~\textit{Fixed Shape}, where parameters are frozen at their initialization; 
(ii)~\textit{Global Learnable}, where a single set of seven parameters is shared across all layers; 
and (iii)~\textit{Layer-wise Learnable}, our proposed configuration. 
The results are detailed in Table~\ref{tab:ablation_granularity}.

\begin{table}[htbp]
\centering
\caption{Parametric Granularity and Learnability on PatternNet Dataset (ResNet-50)}
\label{tab:ablation_granularity}
\resizebox{0.75\columnwidth}{!}{%
\begin{tabular}{lcc}
\toprule
\textbf{Configuration} & \textbf{Learnable Params} & \textbf{OA (\SI{}{\percent})} \\ \midrule
Fixed Soft-ReLU        & 0                         & 99.01            \\
Global Shared          & 7                         & 99.25            \\
\textbf{Layer-wise}    & \textbf{343}        & \textbf{99.67}   \\ \bottomrule
\end{tabular}
}
\end{table}

The study provides two main insights. 
First, the \textit{Layer-wise} approach outperforms the \textit{Global} approach by \SI{0.42}{\percent}, confirming that different feature extraction stages (low-level vs. high-level) benefit from distinct activation behaviors. 
Second, replacing \textit{Fixed} activations with \textit{Learnable} ones improves performance by an additional \SI{0.66}{\percent}, showing that the network can effectively optimize its nonlinearity through backpropagation. 
Importantly, the extra 343 parameters constitute less than \SI{0.002}{\percent} of the total ResNet-50 parameters, so the \textit{Layer-wise} design achieves a noticeable accuracy gain with virtually no computational overhead.

\subsection{Parametric Sensitivity and Geometric Analysis}

The flexibility of the ArcGate formulation is evaluated through a parametric sensitivity analysis, as illustrated in Fig.~\ref{fig:sensitivity}. 
By holding baseline parameters 
constant and varying individual coefficients, we demonstrate how ArcGate navigates diverse functional spaces.
\begin{itemize}
    \item \textbf{Fig.~\ref{fig:sensitivity}(a)~Steepness:} Parameter $a$ modulates the transition rate of the underlying monotonic function. 
    As $a$ increases, the transition from the \emph{off} state to the \emph{on} state becomes more abrupt, allowing the network to learn either gradual or near-binary activation thresholds.
    
    \item \textbf{Fig.~\ref{fig:sensitivity}(b)~Sharpness:} The exponent $p$ regulates the nonlinear stretching of the curvature. Higher values of $p$ result in a sharper \emph{elbow} at the activation point, while lower values ($p \approx 0.5$) yield a smooth, rounded profile similar to self-gated activations like Swish.
    
    \item \textbf{Fig.~\ref{fig:sensitivity}(c)~Horizontal Shift:} The parameter $c$ translates the activation function along the $x$-axis. 
    This functionality serves as a learnable bias at the activation level, enabling the neuron to center its response range on the input distribution.
    
    \item \textbf{Fig.~\ref{fig:sensitivity}(d)~Function Mode:} This plot highlights the bifurcation between gating and saturating behaviors. Setting $\alpha=1$ enables the gating mode ($x \, v(x)$), resulting in an unbounded output suitable for deep feature extraction. Conversely, setting $\beta=1$ (with $\alpha=0$) invokes a saturating mode, producing a bounded response similar to the Sigmoid function.
    
    \item \textbf{Fig.~\ref{fig:sensitivity}(e)~Linear Trend:} By adjusting the linear coefficient $\gamma$, the function introduces a non-zero slope across the domain. This is particularly useful for replicating \emph{Leaky} behavior, which ensures that negative inputs still provide a gradient signal, thereby preventing the \emph{dead ReLU} phenomenon during backpropagation.
    
    \item \textbf{Fig.~\ref{fig:sensitivity}(f)~Unification Demonstration:}   The final sub-figure demonstrates the adaptive nature of ArcGate. By selecting specific parameter sets, the formulation replicates the profiles of standard activation functions, including ReLU, Sigmoid, Tanh, SiLU, and GELU, demonstrating that these disparate functions are special cases of the ArcGate framework.
\end{itemize}

\section{Conclusion}
\label{sec:conclusion}

We introduced the Adaptive Arctangent Gated Activation (ArcGate), which autonomously optimizes its nonlinear profile according to network depth and data characteristics. Evaluated on PatternNet, UC Merced, and EuroSAT MSI, ArcGate consistently outperformed fixed-shape activations, achieving a state-of-the-art \SI{99.67}{\percent} accuracy on PatternNet. It also showed strong structural resilience, retaining a \SI{26.65}{\percent} advantage over ReLU under moderate noise. Analysis of the learned parameters revealed a hierarchical adaptation, with gating strength increasing with depth to improve signal propagation. These results establish ArcGate as a robust and adaptive alternative for high-resolution Earth observation tasks. Future work will investigate its use in lightweight architectures and complex-valued networks for SAR image analysis.


\typeout{get arXiv to do 4 passes: Label(s) may have changed. Rerun}

\end{document}